 \documentclass[final,5p,times,twocolumn,numbers,comma]{elsarticle}
 
\usepackage{amssymb}
\usepackage{amsmath,amsfonts,array}
\usepackage{caption}
\usepackage{lipsum}
\usepackage{float}
\usepackage{hyperref}
\hypersetup{colorlinks=true, citecolor=blue, linkcolor=blue, urlcolor=blue}
\usepackage{cleveref}

\journal{Computer Methods and Programs in Biomedicine}

\begin{document}

\begin{frontmatter}



\title{Drug Synergy Prediction via Residual Graph Isomorphism Networks and Attention Mechanisms}


\author[first]{Jiyan Song \fnref{equal}}
\ead{20221018262@stu.shzu.edu.cn}
\author[first]{Wenyang Wang \fnref{equal}}
\ead{wwyang1213@163.com}
\author[second]{Chengcheng Yan}
\ead{ycc956176796@gmail.com}
\author[first]{Zhiquan Han}
\ead{hanzhiquan@shzu.edu.cn}
\author[first,second]{Feifei Zhao\corref{corresponding}}
\ead{zhaofeifei@shzu.edu.cn}
\cortext[corresponding]{Corresponding author.}
\fntext[equal]{The two authors contribute equally to this work.}

\affiliation[first]{organization={College of Sciences, Shihezi University},
            addressline={221 North 4th Road},
            city={Shihezi},
            postcode={832003}, 
            state={Xinjiang},
            country={China}}
\affiliation[second]{organization={School of Mathematics and Computational Science, Xiangtan University},
            addressline={North 2nd Ring Road},
            city={Xiangtan},
            postcode={411105}, 
            state={Hunan},
            country={China}}

\begin{abstract}
In the treatment of complex diseases, treatment regimens using a single drug often yield limited efficacy and can lead to drug resistance. In contrast, combination drug therapies can significantly improve therapeutic outcomes through synergistic effects. However, experimentally validating all possible drug combinations is prohibitively expensive, underscoring the critical need for efficient computational prediction methods. Although existing approaches based on deep learning and graph neural networks (GNNs) have made considerable progress,  challenges remain in reducing structural bias, improving generalization capability, and enhancing model interpretability. To address these limitations, this paper proposes a collaborative prediction graph neural network that integrates molecular structural features and cell-line genomic profiles with drug-drug interactions to enhance the prediction of synergistic effects. We introduce a novel model named the Residual Graph Isomorphism Network integrated with an Attention mechanism (ResGIN-Att). The model first extracts multi scale topological features of drug molecules using a residual graph isomorphism network, where residual connections help mitigate over-smoothing in deep layers. Subsequently, an adaptive Long Short-Term Memory (LSTM) module fuses structural information from local to global scales. Finally, a cross-attention module is designed to explicitly model drug-drug interactions and identify key chemical substructures. Extensive experiments on five public benchmark datasets demonstrate that ResGIN-Att achieves competitive performance, comparing favorably against key baseline methods while exhibiting promising generalization capability and robustness.
\end{abstract}




\end{frontmatter}




\section{Introduction}
\label{introduction}

The treatment of complex diseases such as cancer and infectious diseases often faces challenges, including limited efficacy of single-drug therapies and the tendency to develop drug resistance \cite{giles2014efficacy, zheng2018drug}. In cancer, multiple mechanisms within cells often undergo changes. Therefore, the traditional treatment model of a single drug targeting a single disease and a single target is often ineffective \cite{wang2023attensyn}. In contrast, combination drug therapy significantly improves efficacy, reduces toxicity that is dependent on drug dosage, and delays the development of drug resistance by using two or more drugs simultaneously \cite{fitzgerald2006systems}. However, drugs may exhibit synergistic, additive, or antagonistic effects in themselves. Using a purely experimental approach to verify all possible combinations is impractical in terms of manpower, time, and resources. This makes computational prediction of drug combinations an essential approach to accelerate the development of combination therapies.

Computational models are widely used to predict and prioritize drug combinations exhibiting synergistic effects. This approach overcomes the limitations of single-agent therapy \cite{al-lazikani2012combinatorial, zhou2024transvae}. The research trajectory in this field is closely intertwined with the development of computational biology and artificial intelligence. It profoundly reflects a paradigm shift from reliance on prior knowledge to data driven approaches, and from shallow statistical learning to deep representational learning \cite{eraslan2019deep}. Early studies primarily relied on prior biological knowledge, including pathway-based rational design \cite{keith2005multicomponent}, network pharmacology \cite{jia2009mechanisms}, and dynamical systems modeling \cite{kang2013inference, van2011systems}. While these approaches established valuable conceptual foundations, their strong dependency on complete and accurate pathway information limited their ability to discover novel interaction patterns beyond known biology. This reliance on predefined biological knowledge restricts model generalization and scalability. Our ResGIN-Att addresses this by learning molecular representations directly from graph structures, requiring no prior pathway knowledge.

The development of high-throughput drug combination screening databases has facilitated the application of data-driven machine learning approaches. Early studies demonstrated the feasibility of computational synergy prediction using artificial neural networks \cite{pivetta2013development} and probabilistic ensemble frameworks \cite{li2015largescale}. Subsequent work employed gradient boosting \cite{josephd2018explainable}, network embedding \cite{liu2019predicting}, support vector machines \cite{shi2019predicting}, and matrix factorization \cite{julkunen2020leveraging} to further improve predictive performance. Among these, comboFM \cite{julkunen2020leveraging} introduced dose–response surface modeling, representing a notable advance in leveraging multi-way interactions. Despite these progresses, traditional machine learning methods remain heavily reliant on handcrafted molecular descriptors and pathway features \cite{lecun2015deep}, and models such as SVM and random forests struggle to effectively process complex unstructured data like molecular graphs \cite{zhang2017from}. This manual feature engineering process is not only labor-intensive but also risks omitting critical structural information. In contrast, our ResGIN-Att performs end-to-end feature learning directly from raw molecular graphs, capturing fine-grained topological patterns without human intervention.

Recent advances in deep learning, combined with the increasing availability of large-scale, high-quality drug combination datasets, have substantially enhanced the accuracy and reliability of computational methods for predicting synergistic drug pairs. Early deep learning models such as DeepSynergy \cite{preuer2018deepsynergy} demonstrated the feasibility of synergy prediction using feedforward networks, but relied on predefined molecular descriptors. Subsequent studies introduced graph convolutional networks \cite{jiang2020deep}  and attention mechanisms \cite{liu2021transynergy} to better capture structural information and long-range dependencies. More recently, end-to-end architectures \cite{liu2024synergnet} and ranking-oriented frameworks \cite{kuru2021matchmaker}, and advanced feature extraction methods \cite{su2022srdfm} have further improved performance. Despite these advances, most existing models still lack explicit modeling of drug–drug interactions and offer limited interpretability \cite{li2023snrmpacdc, xiong2024improving, sun2022deep}. This absence of explicit interaction modeling makes it difficult to understand which substructures contribute to synergy. Our ResGIN-Att addresses this through a cross-attention mechanism that explicitly models mutual influence between drug pairs and identifies key chemical substructures, enhancing both predictive performance and interpretability.

Building upon the widespread adoption of deep learning techniques, research paradigms in computational biomedicine have increasingly shifted toward graph neural networks. GraphSynergy \cite{yang2021graphsynergy} pioneered this direction by constructing drug-pair subnets in PPI networks. Subsequent work integrated attention mechanisms \cite{xiong2024improving}, multi-head attention \cite{yang2021graphsynergy}, knowledge graphs \cite{wang2022deepdds}, and cancer-specific heterogeneous data \cite{hu2022dtsyn} to further improve predictive performance. Despite these advances, existing graph neural network (GNN)-based methods still face critical challenges, including reliance on predefined biological network structures, limited cross-dataset generalization, and insufficient interpretability \cite{li2023snrmpacdc, xiong2024improving, yang2021graphsynergy, wang2022deepdds, hu2022dtsyn}. These structural dependencies hinder model transferability across different biological contexts. Our ResGIN-Att mitigates structural bias via residual connections, enhances cross-dataset generalization through LSTM-based multi-scale feature fusion, and improves interpretability with cross-attention, offering a more generalizable solution.

To address the aforementioned challenges, we propose ResGIN-Att, a deep graph neural network architecture designed to enhance representation learning for drug combination synergy prediction. This model integrates residual graph isomorphism networks, sequential LSTMs, and a cross-attention mechanism to overcome the limitations of existing approaches, including reliance on predefined biological network structures, limited cross-dataset generalization, and insufficient interpretability. The main contributions of this paper are threefold:
\begin{itemize}
    \item We propose a deep residual graph isomorphism network framework for drug synergy prediction, named ResGIN. Its core module integrates residual connections with graph isomorphism networks (GIN) for drug feature learning.
    \item We present a collaborative prediction model that integrates multimodal features and drug interactions. The model employs a novel hierarchical architecture, combining three core modules: ResGIN, sequence LSTM fusion, and cross-attention interaction.
    \item Through comprehensive experiments on five public benchmark datasets, we demonstrate that our ResGIN-Att model demonstrates competitive performance against existing mainstream methods in terms of prediction accuracy, generalization capability, and over-smoothing analysis.
\end{itemize}

\section{Related Work}
As noted in the Introduction, computationally driven drug combination prediction plays a pivotal role in mitigating the inherent limitations of monotherapy. The research paradigm in this field has evolved from reasoning that is driven by theoretical principles, to traditional machine learning based on handcrafted features, and finally to deep learning capable of automatically learning features. The following sections will review three closely related research strands: deep learning for drug combination prediction, applications of GNNs in drug discovery, and the residual connections in GNNs and their applications. This review aims to clarify the technical context and theoretical foundation of this work.
\subsection{Deep Learning for Drug Combination Prediction}
Over the past decade, deep learning methodologies have achieved substantial advances in the computational prediction of anticancer drug combinations \cite{preuer2018deepsynergy, liu2021transynergy, kuru2021matchmaker}. Since DeepSynergy \cite{preuer2018deepsynergy} pioneered the use of feedforward neural networks to integrate chemical descriptors and gene expression profiles in 2018, subsequent deep learning models \cite{jiang2020deep, liu2021transynergy, he2016deep} have progressively shifted towards architectures capable of capturing more complex structured biological data. Studies indicate that prediction performance based on deep learning models has surpassed traditional machine learning approaches \cite{josephd2018explainable, preuer2018deepsynergy, liu2021transynergy}. Furthermore, the deep learning paradigm significantly reduces reliance on manual feature engineering \cite{lecun2015deep, zhang2017from, bai2025predicting}, enabling end-to-end learning from raw or only minimally processed data. This data-driven approach not only enhances predictive performance, but also demonstrates superior modeling capabilities for nonlinear, high dimensional interactions within pharmacological data. However, despite their strong performance, these deep learning methods typically rely on predefined features such as molecular descriptors as inputs, failing to learn directly from the native representation of a molecule's graph structure.

\subsection{Applications of GNNs in Drug Discovery}
GNNs have been increasingly adopted for drug combination prediction, leveraging their capacity to model complex molecular and pharmacological interactions. This enables fuller utilization of molecular structural information and better modeling of drug interactions. Their integration has driven further advancements in this domain. In recent years, the application of GNNs in drug discovery has achieved remarkable success \cite{jiang2020deep, yang2021graphsynergy, wang2022deepdds}. Among these, GraphSynergy \cite{yang2021graphsynergy} innovatively proposed the concept of constructing drug subnetworks on protein-protein interaction networks, utilizing biological networks as the core graph structure for synergistic drug combination prediction, thereby advancing GNN applications in this domain. Subsequent studies \cite{liu2024synergnet, wang2022deepdds, hu2022dtsyn, zhang2023kgansynergy}  have increasingly shifted toward leveraging graph structures to characterize drugs or biological systems with greater precision.

Among numerous GNN architectures, GIN exhibits distinctive representational power. Its design is grounded in the Weisfeiler–Lehman graph isomorphism test, and its learned representations have been theoretically and empirically shown to possess greater discriminative capacity than those of conventional GCN, thereby enabling more precise differentiation of structurally distinct molecular topologies \cite{keyulu2019how}. However, most existing GNN-based drug combination prediction methods still predominantly rely on foundational architectures like GCN or GAT, failing to fully leverage GIN's potential to accurately capture subtle differences in molecular structures. Therefore, in this work, we adopt GIN as the core backbone network, aiming to learn more precise molecular representations through its enhanced expressive power.

\subsection{Residual Connections in GNNs and Applications}
In recent years, residual connections have emerged as a foundational architectural principle in deep neural network design \cite{he2016deep}. Since residual networks achieved breakthrough success in image recognition, subsequent deep models including GNNs \cite{chen2020simple} have widely adopted them as core components for constructing deeper, more powerful networks. Research indicates that incorporating residual connections into GNNs effectively mitigates gradient vanishing and over-smoothing issues that arise with increasing network depth \cite{li2023snrmpacdc, li2024another}. This enables models to aggregate information from broader neighborhoods while preserving detailed features from shallow layers. Furthermore, the residual mechanism significantly reduces the dependency on complex initialization and normalization techniques \cite{xu2018representation}, thereby stabilizing the training process and enhancing model convergence. This architecture not only strengthens the network's representational capabilities but also provides critical assurance for its successful application in complex graph learning tasks such as drug combination prediction.

In summary, deep learning particularly graph convolutional networks (GCNs) has become a pivotal tool for drug combination prediction. However, most existing methods rely on GCN-based architectures. These architectures exhibit limited representational capacity. Consequently, they fail to fully exploit the superior representational advantages offered by graph isomorphism networks. Similarly, when deep networks are constructed to capture more complex dependencies, they have not systematically incorporated key stabilization techniques, such as residual connections. Furthermore, how to explicitly guide models to focus on information on drug-drug interaction  during learning remains an open question. These limitations motivate us to propose a novel model framework.

\section{Methods}

\subsection{Model Architecture}
\begin{figure*}[t]
    \centering
    \includegraphics[width=1\linewidth]{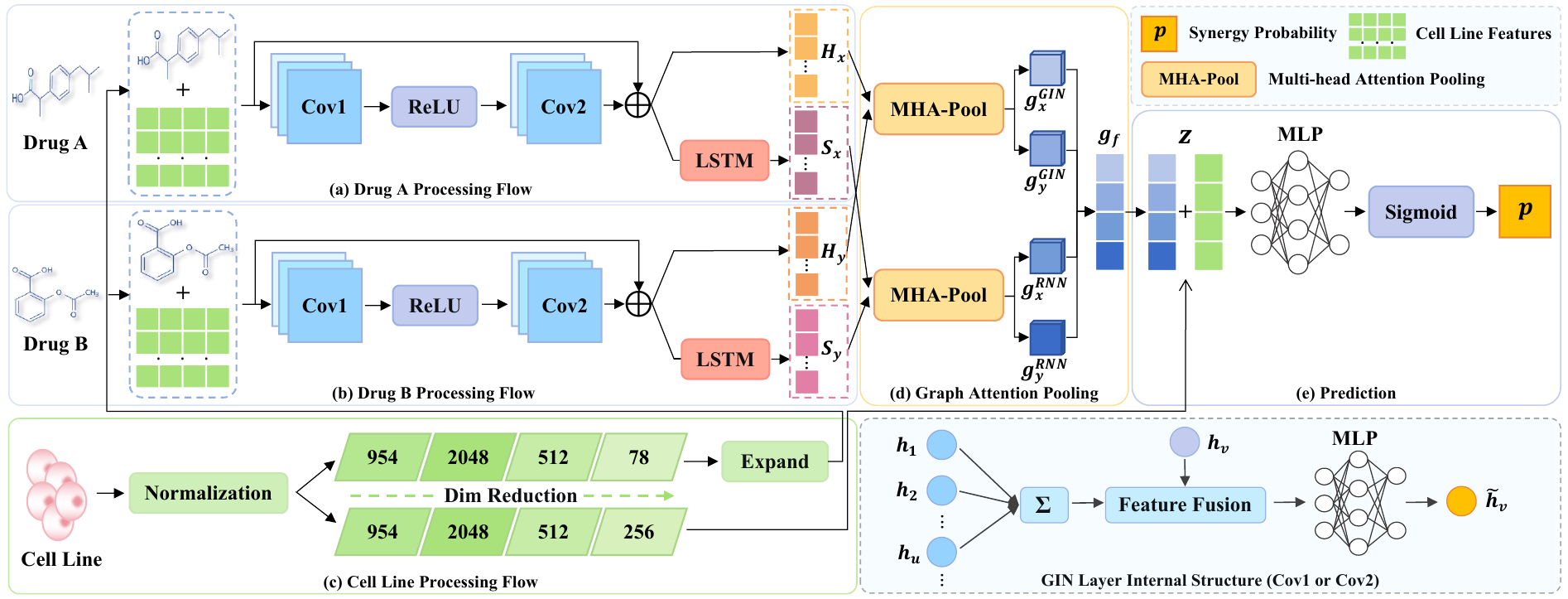}
    \caption{The overall architecture of ResGIN-Att. Given two drugs and their corresponding cell line types, we first convert the SMILES representations of both drugs into molecular graphs featuring initial node attributes and graph structures. Concurrently, cell line attributes obtained from CCLE are added to each node attribute of the drug molecular graph. Next, ResGIN extracts multimodal topological features from the drug molecules. Subsequently, LSTM-based adaptive fusion integrates structural information from local to global levels. A cross-attention module is then designed to explicitly model interactions between drug pairs and identify key chemical substructures. Finally, the prediction module integrates the representations of both drugs with cell line features to forecast synergistic drug combinations.}
    \label{fig:flowchart}
\end{figure*}

This paper introduces ResGIN-Att, a deep graph neural network architecture that integrates attention mechanisms to enhance the accuracy of drug synergy prediction. The overall architecture of the model is shown in Fig.\ref{fig:flowchart} , primarily consisting of three core modules: (1) Molecular Feature Learning Module based on Residual GIN and Sequence LSTM: This module introduces residual connections between GIN layers and combines LSTM networks to extract multiscale features from molecular graphs. (2) Attention-based interaction information learning module: This module calculates importance weights for each atomic substructure within drug molecules. It computes attention scores mutually between feature pairs of drug molecules to better learn interaction information between drug pairs and enhance model interpretability. (3) Prediction Module: This module concatenates drug pair representations with cell line features. The fused features are then fed into a MLP to predict the synergistic effects of the drug combination on a specific cell line.

\subsection{Molecular Feature Learning Based on Residual GIN and Sequence LSTM}
Accurate prediction of drug synergistic effects critically depends on learning highly discriminative, multiscale feature representations from molecular structures of candidate drugs. To overcome the limitations of traditional GCNs in distinguishing complex molecular topologies and to effectively capture information at multiple levels, ranging from details at the atomic level to patterns at the molecular level, this work designs a molecular feature learning module that integrates ResGIN with sequence LSTM networks.

\subsubsection{Molecular Graph Construction and Feature Initialization}

Given the SMILES representation of a drug compound, we employ the open-source cheminformatics toolkit RDKit \cite{landrum2013rdkit} to construct its corresponding molecular graph $\mathbf{G = (V, E)}$, where $\mathbf{V}$ denotes the set of atomic nodes and $\mathbf{E}$ denotes the set of covalent bonds. For each node $v_i \in \mathbf{V}$, we extract its fundamental feature vector $x_i \in \mathbb{R}^{d_a}$ using the feature encoding function $\phi_{\text{atom}}$:
\begin{equation}
x_i = \phi_{\text{atom}}(v_i),
\label{eq:feature_encoding_phi}
\end{equation}
where $\mathbb{R}^{d_a}$ denotes a real vector space of dimension $d_{a}$. This vector $x_{i}$ is the feature representation of node $v_{i}$, comprising $d_{a}$ real-valued features describing the node's attributes.

For each cancer cell line, we extract its gene expression profile $x_{cl} \in \mathbb{R}^{d_g}$ from the Cancer Cell Line Encyclopedia (CCLE), where $\mathbb{R}^{d_g}$ denotes a real vector space of dimension $d_{g}$, and $d_g$ represents the feature dimension of the gene expression profile. To integrate these profiles with atomic-level features, we employ a dedicated MLP for dimensionality reduction and nonlinear transformation to adapt it to the atomic feature space:
\begin{equation}
c = \mathrm {MLP}_{\text{c}}(r_{\text{cl}}), \quad c \in \mathbb{R}^{d_{a}}.
\end{equation}

Through additive operations, we fuse the transformed cell line feature vectors with the basis feature vectors of each atom, ultimately forming the initial representation of the final node $h_i^{0}$:
\begin{equation}
{h}_{i}^{0} = x_{i} + c.
\end{equation}
\subsubsection{Residual Graph Isomorphism Network}
In molecular graph representation learning, the core of GNNs lies in updating node representations through neighborhood aggregation. However, many common GNN variants are bound in their representational power, which limits their ability to distinguish between certain distinct topological structures. In other words, they are not robust to the ``graph isomorphism test". To address this, we adopt the Graph Isomorphism Network. Its design strictly follows the Weisfeiler-Lehman graph isomorphism test framework. This proves that it is one of the most representational architectures within the GNN paradigm.

The standard GIN layer updates node representations via a learnable, injective aggregation function, thereby guaranteeing strong structural discriminability. For a node $v$, its embedding vector $h_v^{(k)}$ in layer $k$ is defined as:
{
\small
\begin{equation}
h_v^{(k)} = \mathrm{MLP}^{(k)}\left((1 + \epsilon^{(k)}) \cdot h_v^{(k-1)} + \sum_{u \in {N}(v)} h_u^{(k-1)}\right),
\end{equation}
}where $N(v)$ denotes the set of direct neighbors of node $v$, $\epsilon^{(k)}$ is a learnable scalar parameter used to adjust the importance of a node's own features in the aggregation process. $u$ is a neighbor node of $N(v)$. $\mathrm{MLP}^{(k)}$ is a multilayer perceptron applied to each node, endowing the model with powerful nonlinear feature transformation capabilities.

Despite the strong representational capacity of GINs, deep GNNs remain susceptible to over-smoothing and vanishing gradient phenomena. As the number of network layers increases, the receptive fields of nodes continuously expand, causing features from different nodes to converge toward the same value and thereby losing their unique structural information. To address this issue and build deeper, more powerful molecular representation networks, we introduce residual connections into GNNs, constructing ResGIN modules.

First, compute the raw output $\tilde{h}_v^{(k)}$ of the GIN layer:
{
\small
\begin{equation}
\tilde{h}_v^{(k)} = \mathrm{MLP}^{(k)}\left((1 + \epsilon^{(k)})\cdot{h}_v^{(k-1)} + \sum_{u \in \mathcal{N}(v)}{h}_u^{(k-1)}\right).
\label{eq:gin_update}
\end{equation}
}Subsequently, through a skip connection, the input $h_v^{(k-1)}$ from the previous layer is added to the original output of this layer to obtain the final output of this layer $h_v^{(k)}$:

\begin{equation}
h_v^{(k)} = h_v^{(k-1)} + \tilde{h}_v^{(k)}.
\end{equation}

After stacking ${k}$ layers of the ResGIN architecture, we obtain the final node representation $h_i^{(k)}$. To facilitate efficient tensor computations in subsequent attention pooling modules, we organize the embeddings of all nodes into a node feature matrix as the final output of this module. Specifically, we stack the feature vectors of all nodes row-wise to form the graph-level node feature matrix $H \in \mathbb{R}^{N\times d_h}$  for this drug:
\begin{equation}
H=[h_1^{(k)}, h_2^{(k)}, ..., h_N^{(k)}]^{T},
\end{equation}
where $N$ denotes the total number of nodes in the molecular graph, and $d_h$ represents the embedding dimension of the node. Each row $H[i,:]$ of matrix $H$ corresponds to the transposed eigenvector $(h_i^{(k)})^T$ of node $i$.
\subsubsection{Multimodal Feature Fusion Based on LSTM}
The ResGIN module outputs node representations that encode multi-level structural information. Among these, shallow features $H^{(1)}$ encode the local chemical environment, while deep features $H^{(k)}$ integrate more global molecular topological information. To adaptively fuse these multi-resolution features that span from local to global scales, inspired by MR-GNN \cite{heaton2018ian}, we introduce a LSTM network to model this sequence of feature.

For each node $v_i$ in the graph, we treat its features $[h_i^{(1)}, h_i^{(2)}, ..., h_i^{(K)}]$ across all GIN levels as a time series of length $K$. The LSTM processes this sequence sequentially, with the update process as follows:
\begin{equation}
s_i^{(l)} = \text{LSTM}(s_i^{(l-1)}, h_i^{(l)}), \quad l = 1, 2, \ldots, K,
\end{equation}
where $h_i^{(l)}$ is the output of node $v_i$ at layer $l$ of ResGIN, serving as the input to the LSTM at time step $l$. $s_i^{(l)}$ is the hidden state of the LSTM after processing the features at layer $l$, which integrates feature information from layers 1 to $l$. The initial hidden state $s_i^{(0)}$ of the LSTM is initialized as the zero vector.

Upon completion of sequence processing, the final hidden state $s_i^{(K)}$ serves as an enhanced node representation for $v_i$, encoding multi-resolution structural information. Performing this operation on all nodes produces the LSTM-fused node feature matrix $S=[s_1^{(K)},s_2^{(K)},...,s_N^{(K)}]^{T}\in\mathbb{R}^{N\times d_s}$ throughout the graph, where $d_s$ denotes the dimension of the LSTM hidden layer and $N$ represents the total number of nodes in the molecular graph.
\subsection{Cross-Attention Mechanism}
The aforementioned module yields rich intrinsic characterizations of drug molecules. However, accurate prediction of drug synergistic effects depends not only on the intrinsic properties of individual drugs but also on the pairwise molecular interactions between them. To capture this complex pairwise interaction information while simultaneously identifying key chemical substructures that significantly contribute to synergistic effects, interpretability, we designed an interaction information learning module based on an attention mechanism. At its core lies a cross-attention pooling mechanism that generates a set of importance weights reflecting the mutual influence between drug pairs (Drug$A$, Drug$B$). This process yields an enhanced graph-level representation. We first process the features from both paths separately. For the ResGIN path, we use the characteristic matrices of the nodes $H_x^{(K)}$ and $H_y^{(K)}$ of the final layer. For the LSTM path, we use the fused node feature matrices $S_x$ and $S_y$. Next, we will illustrate the process using the features $H_x$ and $H_y$ of the ResGIN path as an example. The specific calculation steps are as follows:

First, compute the pairwise interaction attention score matrices $A_x$ and $A_y$ for the drug pair:
\begin{align}
A_x &= \tanh(H_x W_q \cdot (H_y W_k)^T), \\
A_y &= \tanh(H_y W_q \cdot (H_x W_k)^T),
\end{align}
where $H_x\in\mathbb{R}^{N_x\times d_h}$ and $H_y\in \mathbb{R}^{N_y\times d_h}$ represent the characteristic matrices of the nodes for drug A and drug B. $W_q, W_k\in \mathbb{R}^{d_h\times d_a}$ are trainable weight matrices used to project node features onto a common $d_a$ dimensional attention space. $A_x \in \mathbb{R}^{N_x\times N_y}$, whose elements $(A_{x})_{i,j}$ quantify the interaction strength between the $i^{th}$ substructure in drug $A$ and the $j^{th}$ substructure in drug $B$, $A_y$ follows similarly.

Next, the attention scores are aggregated along the dimension corresponding to the partner drug and subsequently normalized to yield importance weights for each drug node:
\begin{align}
a_x &= \text{softmax}\left(\sum_{j=1}^{N_y} (A_{x})_{i,j}\right),\\
a_y &= \text{softmax}\left(\sum_{i=1}^{N_x} (A_{y})_{j,i}\right),
\end{align}
where $a_x \in \mathbb{R}^{Nx}$ and $a_y \in \mathbb{R}^{Ny}$ represent the node importance vectors for drug $A$ and drug $B$, respectively. Each element within these vectors indicates the relative importance of the corresponding node within the drug combination.

Finally, the weights of importance computed are applied to derive the graph-level representations enhanced by interaction $g_x$ and $g_y$:
\begin{align}
g_x &= \sum_{i=1}^{N_x} a_x^{(i)} \cdot (H_x W_v)^{(i)},\\
g_y &= \sum_{i=1}^{N_y} a_y^{(i)} \cdot (H_y W_v)^{(i)},
\end{align}
where $W_v$ represents the shared weight.

By performing the exact same processing steps on the $S_x$ and $S_y$ features of the LSTM path, another pair of layer-level representations $g_x^{LSTM}$ and $g_y^{LSTM}$ can be obtained:
\begin{align}
g_{x}^{\text{LSTM}},\ g_{y}^{\text{LSTM}} = \text{AP}(S_{x}, S_{y}),
\end{align}
where AP denotes the aforementioned attention-based pooling operation.

The final output of this module is the concatenation of four graph-level representations: two derived from the graph pathway and two from the sequence (LSTM) pathway. Assuming each graph-level representation vector $g\in\mathbb{R}^{d}$, the concatenated final vector $g_f$ is as follows:
\begin{align}
g_{f} &= g_{x}^{\text{GIN}} \parallel g_{y}^{\text{GIN}} \parallel g_{x}^{\text{LSTM}} \parallel g_{y}^{\text{LSTM}} \in \mathbb{R}^{4\times d},
\label{eq:final_feature_align}
\end{align}
where the symbol $\parallel$ denotes the concatenation operation of vectors along the dimension of the feature.
\subsection{Prediction Module}
The prediction module receives the fused representation $g_{f}$ from the attention interaction module, concatenates it with the projected cell-line features, and constructs the final feature vector $z$:
\begin{align}
z &= g_{f}\parallel{\text{MLP}_c(r_{\text{cl}})}.
\end{align}

Subsequently, $z$ is input into a multi-layer perceptron to obtain the final prediction output $p$:
\begin{align}
p &= \sigma\left( \text{MLP}_{\text{pred}}(z)\right),
\end{align}
for binary classification tasks, $\sigma$ represents the sigmoid function, where the output $p \in (0,1)$ indicates the predicted frequency of synergistic effects in drug combinations.

For binary classification tasks, the sigmoid function is employed instead of the softmax function to improve computational efficiency. Sigmoid directly outputs the probability of the positive class $p \in (0,1)$, whereas Softmax requires computing probabilities for both classes and normalizing them to sum to 1. Furthermore, it is inherently compatible with the binary cross-entropy loss function, resulting in stable gradient descent and facilitating model optimization. This makes it the standard approach for such tasks.

Our model training optimizes all parameters by minimizing the cross-entropy loss function $\mathcal{L}$, defined as:
\begin{equation}
\mathcal{L} = -\frac{1}{N} \sum_{i=1}^{N} \left[ y_{i} \log \left( p_{i} \right) + \left( 1-y_{i} \right) \log \left( 1-p_{i} \right) \right],
\label{eq:binary_cross_entropy}
\end{equation}
where $N$ denotes the total number of training samples, $y_i \in {(0, 1)}$ represents the true label of the sample $i$, and $p_i$ is the probability predicted by the model that sample $i$ is collaborative.
\section{Experiments and Results}
To comprehensively evaluate the effectiveness, generalization capability, and robustness of our proposed residual GIN-based model with integrated attention mechanisms, we conduct systematic experiments. These experiments aim to address the following key questions: (1) Does the model maintain excellent performance across diverse datasets of varying scales and origins? (2) How does the model perform in predicting synergistic effects of anticancer drugs compared to existing mainstream methods? (3) How do the key components within the model individually contribute to its overall performance?

To this end, we perform comprehensive evaluations on five widely adopted public benchmark datasets, comparing our method against a diverse set of baseline approaches. We also performed ablation studies to validate the rationality of our model architecture, analyzed parameter sensitivity to assess model robustness, and further investigated the over-smoothing phenomenon through layer-wise sensitivity analysis. This section details the experimental setup, analyzes the comparative results, and discusses the key research findings.

\subsection{Experimental Setup}
\subsubsection{Datasets}
We evaluate our method on five publicly available benchmark datasets. \Cref{tab:dataset_stats} shows the statistics of these datasets.
\begin{table}[!t]
\centering
\caption{Dataset Statistics}
\label{tab:dataset_stats}
\begin{tabular}{>{\centering\arraybackslash}p{2.5cm} >{\centering\arraybackslash}p{1cm} >{\centering\arraybackslash}p{1.5cm} >{\centering\arraybackslash}p{2cm}}
\hline
Dataset & Drugs & Cell Lines & Samples \\
\hline
O'Neil  & 38 & 39 & 23,062 \\
ALMANAC & 118 & 118 & 296,503 \\
Oncology Screen & 21 & 29 & 4,176 \\
DrugCombDB & 600 & 68 & 60,932 \\
DrugComb & 354 & 170 & 330,917 \\
\hline
\end{tabular}
\end{table}

\subsubsection{Baseline Methods}
To comprehensively evaluate the performance of our model, we compare it against a representative set of state-of-the-art methods spanning diverse technical paradigms, including graph neural networks, multimodal learning, and conventional machine learning approaches:

AttenSyn \cite{wang2023attensyn} employs a hierarchical attention mechanism to learn deep interactions between drugs and gene expression profiles of cell lines. Its core lies in aggregating cell line features through attention that is aware of the drugs and aggregating drug features through attention that is aware of the cell lines, thereby effectively capturing key signals within multimodal information.

DTSyn \cite{jiang2020deep} delves into the mechanisms of drug synergism within specific cell line contexts. It constructs a heterogeneous network comprising relationships between drugs and targets and those between diseases and genes, learns representations in a low dimensional space of drugs and cell lines through network embedding techniques, and ultimately employs deep neural networks to predict synergistic probabilities. Its strength lies in integrating the prior knowledge of the rich biological network.

MR-GNN \cite{xu2019mrgnn} employs a multi-relational GNN to model drug chemical structures. MR-GNN treats molecular graphs as multi-relational graphs, leveraging graph convolutional networks to learn information conveyed by different chemical bond types between atoms. This approach extracts more expressive drug representations that are then combined with cell line features for synergistic prediction.

DeepSynergy \cite{preuer2018deepsynergy} is one of the earliest and classic approaches to applying deep learning in this field. DeepSynergy employs a simple perceptron with multiple layers, directly concatenating drug chemical descriptors with cell line gene expression profiles as input. Despite its relatively straightforward model architecture, it laid the foundation for drug synergy prediction using deep learning and remains a crucial comparative benchmark to this day.
\subsubsection{Evaluation Metrics}
To comprehensively evaluate model performance, we adopt a multidimensional evaluation metric suite. All metrics are defined based on four core statistics in the binary classification confusion matrix: True Positive (TP) represents the number of positive cases correctly predicted by the model. True Negative (TN) represents the number of negative cases correctly predicted by the model. False Positive (FP) represents the number of cases that are actually negative but are incorrectly classified as positive by the model. False Negative (FN) represents the number of cases that are actually positive but incorrectly classified as negative by the model. Based on the above statistical quantities, Accuracy (ACC) measures the overall classification precision. Precision (PREC) and Recall (RECALL) respectively, assess the accuracy of predicting positive instances and the coverage of true positives. True Positive Rate (TPR) and True Negative Rate (TNR) reflect the model's ability to identify positive and negative instances, respectively. Furthermore, to mitigate evaluation bias caused by class imbalance, we specifically employ balanced accuracy (BACC) as a key metric to assess model equilibrium performance, which is defined as the arithmetic mean of TPR and TNR. The calculation formulas for each metric are as follows:
\begin{equation}
\left\{
    \begin{array}{l}
        \mathit{ACC}    = \dfrac{\mathit{TP}+\mathit{TN}}{\mathit{TP}+\mathit{TN}+\mathit{FP}+\mathit{FN}},\\[8pt]
        \mathit{PREC}   = \dfrac{\mathit{TP}}{\mathit{TP}+\mathit{FP}},\\[8pt]
        \mathit{RECALL} = \dfrac{\mathit{TP}}{\mathit{TP}+\mathit{FN}},\\[8pt]
        \mathit{TPR}    = \dfrac{\mathit{TP}}{\mathit{TP}+\mathit{FN}},\\[8pt]
        \mathit{TNR}    = \dfrac{\mathit{TN}}{\mathit{TN}+\mathit{FN}},\\[8pt]
        \mathit{BACC}   = \dfrac{\mathit{TPR}+\mathit{TNR}}{2}.
    \end{array}
\right.
\label{eq:metrics}
\end{equation}

\subsubsection{Implementation Details}

\textbf{Data Partitioning Strategy:} To comprehensively evaluate the model’s generalization capability and mitigate bias introduced by random data partitioning, we employ five-fold cross-validation in all experiments. Specifically, the entire dataset is randomly divided into five mutually exclusive subsets. In each experimental iteration, one subset is designated as the test set, while the remaining four serve as the training set. The final performance metric is calculated as the average and standard deviation of the five test results. This strategy ensures that all samples participate in testing exactly once, yielding a more robust estimate of model performance.

\textbf{Hyperparameter Tuning:} The model hyperparameters were determined via a systematic procedure combining preliminary empirical analysis and grid search.  \Cref{tab:hyperparameter_settings} shows the  the hyperparameter settings.

\begin{table}[!t]
\centering
\caption{Hyperparameter Settings.}
\label{tab:hyperparameter_settings}
{\small
\begin{tabular*}{\linewidth}{@{\extracolsep{\fill}} c c c}
\hline
\multicolumn{3}{c}{Model Architecture Parameters} \\
\hline
Parameters & Value & Instruction \\
\hline
molecule\_channels & 78 & Molecular feature dimension \\
hidden\_channels   & 128 & Hidden layer dimension \\
middle\_channels   & 64  & Intermediate layer dimension \\
layer\_count       & 2   & Number of GIN layer \\
out\_channels      & 2   & Number of output classes \\
num\_heads         & 4   & Number of attention heads \\
train\_batch\_size & 128 & Training batch size \\
test\_batch\_size  & 128 & Testing batch size \\
lr              & 0.0005 & Learning rate \\
num\_epochs      & 200 & Number of training epochs \\
n\_folds           & 5   & Number of cross-validation folds \\
\hline
\end{tabular*}
}
\end{table}

\textbf{Training Environment:} All experiments were conducted in a standardized hardware and software environment to ensure result reproducibility. This experiment was conducted on a computing platform configured with an Intel (R) Core (TM) i5-1240P processor, an NVIDIA RTX 3060 graphics card (12GB VRAM), and 70GB of memory. The system operated in a CUDA 11.2 environment, with a 20GB system disk and a 50GB data disk. The software environment utilized Python 3.9 and the PyTorch 2.2.0 framework.
\subsection{Main Results}
\subsubsection{Evaluation of Generalization Performance Across Multiple Datasets}
To validate the generalization capability and robustness of our proposed model, we conduct a systematic evaluation across five widely adopted public benchmark datasets. 

Our model achieves the best performance on the widely adopted O’Neil benchmark dataset, attaining top-rank results across all six evaluation metrics. In particular, it achieved an AUC of 0.921 for the comprehensive discriminative capacity metric, while also achieving an ACC and BACC of 0.840. This fully demonstrates the effectiveness of the model architecture in capturing key features of drug synergistic interactions. The model maintained high performance across the remaining four datasets, with AUC values consistently ranging between 0.873 and 0.912. This outcome indicates that the feature representations and prediction mechanisms learned by our model are not overfitted to a single data distribution. Instead, they demonstrate robust cross data set generalization capabilities, adapting effectively to variations arising from different experimental conditions and data sources.

Across all benchmark datasets, our model demonstrates balanced performance in PREC and RECALL (two metrics critical to model utility). Taking the O'Neil dataset as an example, both PREC and Recall reach 0.829. This indicates that the model identifies as many genuine synergistic drug combinations as possible while ensuring high confidence in the predicted synergistic pairs. This equilibrium is vital for guiding real biomedical experiments and preventing resource wastage. The standard deviation of all reported results remains low, reflecting minimal performance fluctuation across different data partitions. This demonstrates the high robustness and reproducibility of the model.

In summary, the experiments presented in this section provide systematic empirical evidence of the strengths of our proposed model. It not only achieves competitive performance on specific datasets but also offers a reliable solution with promising generalization capabilities, contributing to the computational prediction of drug synergistic effects. 

\subsubsection{In-Depth Analysis on a Primary Dataset}
To rigorously assess the performance advantages of our model over existing state-of-the-art methods, we conduct an in-depth comparative analysis. The evaluation was performed on five authoritative benchmark datasets. It included comparisons with multiple representative approaches, such as AttenSyn, DTSyn, MR-GNN, and DeepSynergy. The comparison results are detailed in \Cref{tab:O'Neil performance_comparison,tab:ALMANAC performance_comparison,tab:Oncology_Screen_performance_comparison,tab:DrugCombDB_performance_comparison,tab:DrugComb_performance_comparison}.
\begin{table*}[!t]
\centering
\caption{Performance Comparison of Our Model and Existing Models on O'Neil Dataset.}
\label{tab:O'Neil performance_comparison}
\small
\begin{tabular*}{\linewidth}{@{\extracolsep{\fill}} c c c c c c c}
\hline
\textbf{Models} & \textbf{AUC} & \textbf{ACC} & \textbf{F1} & \textbf{PREC} & \textbf{Recall} & \textbf{BACC} \\
\hline
AttenSyn & 0.915 ± 0.007 & 0.836 ± 0.008 & 0.823 ± 0.011 & 0.829 ± 0.007 & 0.818 ± 0.019 & 0.835 ± 0.008 \\

DTSyn & 0.885 ± 0.008 & 0.806 ± 0.007 & 0.785 ± 0.006 & 0.834 ± 0.015 & 0.795 ± 0.015 & 0.806 ± 0.006 \\

MR-GNN & 0.894 ± 0.010 & 0.813 ± 0.009 & 0.806 ± 0.009 & 0.810 ± 0.020 & 0.807 ± 0.013 & 0.815 ± 0.009 \\

DeepSynergy & 0.715 ± 0.006 & 0.722 ± 0.006 & 0.703 ± 0.008 & 0.731 ± 0.022 & 0.702 ± 0.011 & 0.721 ± 0.005 \\

\textbf{ResGIN-Att} & \textbf{0.921 ± 0.005} & \textbf{0.840 ± 0.009} & \textbf{0.829 ± 0.009} & \textbf{0.829 ± 0.020} & \textbf{0.829 ± 0.016} & \textbf{0.840 ± 0.008} \\
\hline
\end{tabular*}
\end{table*}

\begin{table*}[!t]
\centering
\caption{Performance Comparison of Our Model and Existing Models on ALMANAC Dataset.}
\label{tab:ALMANAC performance_comparison}
\small
\begin{tabular*}{\linewidth}{@{\extracolsep{\fill}} c c c c c c c}
\hline
\textbf{Models} & \textbf{AUC} & \textbf{ACC} & \textbf{F1} & \textbf{PREC} & \textbf{Recall} & \textbf{BACC} \\
\hline
AttenSyn & 0.853 ± 0.008 & 0.794 ± 0.006 & 0.787 ± 0.011 & 0.784 ± 0.009 & 0.773 ± 0.015 & 0.791 ± 0.009 \\

DTSyn & 0.811 ± 0.006 & 0.765 ± 0.007 & 0.759 ± 0.005 & 0.742 ± 0.013 & 0.756 ± 0.013 & 0.733 ± 0.005 \\

MR-GNN & 0.834 ± 0.009 & 0.783 ± 0.008 & 0.771 ± 0.007 & 0.767 ± 0.012 & 0.733 ± 0.011 & 0.767 ± 0.010 \\

DeepSynergy & 0.711 ± 0.005 & 0.705 ± 0.005 & 0.701 ± 0.009 & 0.722 ± 0.018 & 0.700 ± 0.009 & 0.709 ± 0.007 \\

\textbf{ResGIN-Att} & \textbf{0.873 ± 0.006} & \textbf{0.801 ± 0.007} & \textbf{0.795 ± 0.013} & \textbf{0.803 ± 0.015} & \textbf{0.784 ± 0.019} & \textbf{0.794 ± 0.010} \\
\hline
\end{tabular*}
\end{table*}

\begin{table*}[!t]
\centering
\caption{Performance Comparison of Our Model and Existing Models on Oncology Screen Dataset.}
\label{tab:Oncology_Screen_performance_comparison}
\small
\begin{tabular*}{\linewidth}{@{\extracolsep{\fill}} c c c c c c c}
\hline
\textbf{Models} & \textbf{AUC} & \textbf{ACC} & \textbf{F1} & \textbf{PREC} & \textbf{Recall} & \textbf{BACC} \\
\hline
AttenSyn & 0.882 ± 0.006 & 0.801 ± 0.006 & 0.799 ± 0.009 & 0.806 ± 0.006 & 0.801 ± 0.010 & 0.809 ± 0.007 \\

DTSyn & 0.863 ± 0.007 & 0.775 ± 0.010 & 0.765 ± 0.007 & 0.781 ± 0.012 & 0.778 ± 0.010 & 0.781 ± 0.007 \\

MR-GNN & 0.871 ± 0.007 & 0.786 ± 0.006 & 0.783 ± 0.010 & 0.793 ± 0.012 & 0.785 ± 0.009 & 0.809 ± 0.009 \\

DeepSynergy & 0.709 ± 0.005 & 0.715 ± 0.007 & 0.698 ± 0.007 & 0.720 ± 0.018 & 0.691 ± 0.011 & 0.703 ± 0.008 \\

\textbf{ResGIN-Att} & \textbf{0.906 ± 0.009} & \textbf{0.827 ± 0.009} & \textbf{0.816 ± 0.007} & \textbf{0.821 ± 0.017} & \textbf{0.820 ± 0.016} & \textbf{0.834 ± 0.009} \\
\hline
\end{tabular*}
\end{table*}

\begin{table*}[!t]
\centering
\caption{Performance Comparison of Our Model and Existing Models on DrugCombDB Dataset.}
\label{tab:DrugCombDB_performance_comparison}
\small
\begin{tabular*}{\linewidth}{@{\extracolsep{\fill}} c c c c c c c}
\hline
\textbf{Models} & \textbf{AUC} & \textbf{ACC} & \textbf{F1} & \textbf{PREC} & \textbf{Recall} & \textbf{BACC} \\
\hline
AttenSyn & 0.895 ± 0.005 & 0.814 ± 0.008 & 0.803 ± 0.006 & 0.811 ± 0.005 & 0.809 ± 0.008 & 0.827 ± 0.008 \\

DTSyn & 0.862 ± 0.009 & 0.778 ± 0.009 & 0.760 ± 0.009 & 0.803 ± 0.014 & 0.780 ± 0.009 & 0.806 ± 0.008 \\

MR-GNN & 0.879 ± 0.010 & 0.795 ± 0.009 & 0.785 ± 0.009 & 0.805 ± 0.020 & 0.798 ± 0.013 & 0.815 ± 0.009 \\

DeepSynergy & 0.715 ± 0.006 & 0.722 ± 0.006 & 0.703 ± 0.008 & 0.731 ± 0.022 & 0.702 ± 0.011 & 0.721 ± 0.005 \\

\textbf{ResGIN-Att} & \textbf{0.912 ± 0.004} & \textbf{0.829 ± 0.013} & \textbf{0.818 ± 0.005} & \textbf{0.824 ± 0.015} & \textbf{0.820 ± 0.013} & \textbf{0.835 ± 0.005} \\
\hline
\end{tabular*}
\end{table*}

\begin{table*}[!t]
\centering
\caption{Performance Comparison of Our Model and Existing Models on DrugComb Dataset.}
\label{tab:DrugComb_performance_comparison}
\small
\begin{tabular*}{\linewidth}{@{\extracolsep{\fill}} c c c c c c c}
\hline
\textbf{Models} & \textbf{AUC} & \textbf{ACC} & \textbf{F1} & \textbf{PREC} & \textbf{Recall} & \textbf{BACC} \\
\hline
AttenSyn & 0.876 ± 0.009 & 0.806 ± 0.007 & 0.801 ± 0.007 & 0.803 ± 0.006 & 0.795 ± 0.010 & 0.801 ± 0.006 \\

DTSyn & 0.844 ± 0.008 & 0.765 ± 0.010 & 0.755 ± 0.008 & 0.787 ± 0.013 & 0.765 ± 0.013 & 0.782 ± 0.008 \\

MR-GNN & 0.856 ± 0.009 & 0.773 ± 0.007 & 0.785 ± 0.007 & 0.793 ± 0.017 & 0.776 ± 0.012 & 0.787 ± 0.009 \\

DeepSynergy & 0.720 ± 0.008 & 0.729 ± 0.009 & 0.710 ± 0.009 & 0.731 ± 0.022 & 0.710 ± 0.010 & 0.726 ± 0.007 \\

\textbf{ResGIN-Att} & \textbf{0.895 ± 0.007} & \textbf{0.812 ± 0.013} & \textbf{0.809 ± 0.010} & \textbf{0.815 ± 0.017} & \textbf{0.804 ± 0.016} & \textbf{0.815 ± 0.007} \\
\hline
\end{tabular*}
\end{table*}

Our model achieves top-rank performance across all six evaluation metrics on each of the five benchmark datasets. This fully demonstrates the compehensive superiority of our proposed architecture.

Our model demonstrates consistently strong performance across a diverse set of benchmark datasets. For example, its AUC scores across five datasets are as follows: 0.921 (O'Neil), 0.873 (ALMANAC), 0.906 (Oncology-Screen), 0.912 (DrugCombDB) and 0.895 (DrugComb). The small fluctuation range and consistently high values confirm the model's adaptability to diverse data sources and distributions, demonstrating strong generalization capabilities. Additionally, our model  generally exhibits low standard deviations in metrics. For instance, on the DrugCombDB dataset, the standard deviation of our model's AUC is 0.004, lower than AttenSyn's 0.005. This demonstrates superior robustness and reproducibility in our model's predictions.

In terms of PREC and RECALL, a pair of metrics that typically require trade-offs, this model demonstrates superior balance. Taking the Oncology-Screen dataset as an example, our model achieves PREC and Recall of 0.821 and 0.820, respectively, while the next-best performer, AttenSyn, achieves 0.806 and 0.801. This indicates that our model ensures the reliability of predicted synergistic combinations (high precision) while minimizing the omission of genuine synergistic combinations (high recall), thereby offering greater guidance value for biomedical experiments. A comprehensive comparison of F1 scores further corroborates this, with our model achieving the highest F1 score across all datasets.

Compared to graph-based learning methods, including MR-GNN and DTSyn, our model demonstrates higher accuracy in representing molecular structures and drug-drug interactions, a advantage confirmed by improvements in several key evaluation metrics, such as AUC, ACC, and F1 score. Compared to the AttenSyn model, which similarly integrates attention mechanisms with sequence modeling, this work innovatively introduces a residual connection mechanism based on the GIN architecture. This design effectively mitigates the excessive smoothing issue commonly found in deep graph neural networks and enhances the ability to preserve critical local features of nodes, thereby achieving more stable and consistent performance across datasets of varying scales and distributions. Overall, this study further confirms that deep learning, particularly methods based on graph neural networks, possesses strong modeling capabilities and application potential in drug synergy prediction tasks, and provides a more reliable computational tool for drug combination screening aimed at precision medicine.

The comparative analysis presented in this section provides empirical support for the effectiveness of our proposed model. It demonstrates competitive performance not only on individual datasets but also across multiple benchmark datasets with diverse characteristics. The model achieves favorable results in terms of general discriminative capability (AUC), classification accuracy (ACC, BACC), and precision-recall balance (F1). These findings suggest that the proposed approach offers a reliable option for predicting drug synergistic effects.

\subsection{Ablation Study}
To validate the contribution of each key component in the ResGIN-Att architecture, we conduct an ablation study across five benchmark datasets. These experiments investigate the contributions of the residual connection and the GIN architecture to the model's performance. We designed two variants of the ResGIN-Att as follows:
\begin{itemize}
    \item \textbf{w/o GIN:} ResGIN-Att replaces the GIN with a normal GCN.
    \item \textbf{w/o residual connection:} ResGIN-Att removes the residual connection module.
\end{itemize}

\begin{figure*}[htbp]
    \centering
    \includegraphics[width=\textwidth,keepaspectratio]{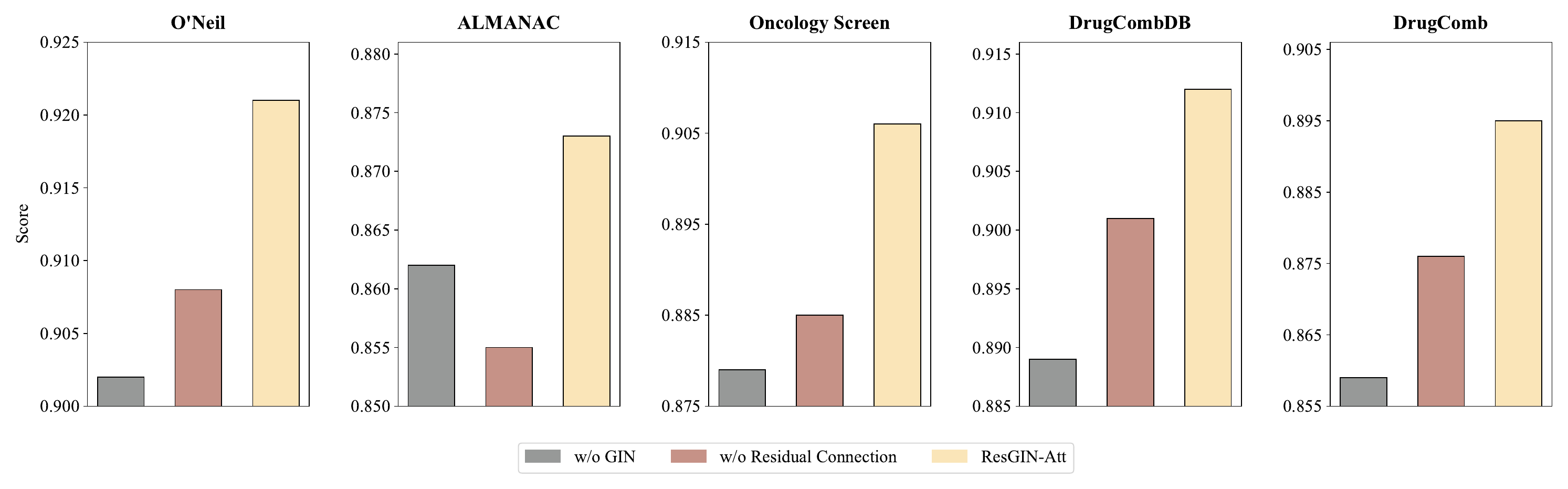}
    \caption{AUC comparison of ablation study on five datasets.}
    \label{fig:ablation_study_auc}
\end{figure*}

The results of the ablation experiments are shown in Fig.\ref{fig:ablation_study_auc}.After replacing the original GNN architecture with GIN, the model performance showed a steady upward trend. GIN's design is based on the Weisfeiler-Lehman graph isomorphism test. This endows it with enhanced structural discriminative capabilities. It can make finer distinctions between different molecular topological structures, thereby learning more discriminative molecular representations. Experimental data show that adding residual connections to the AttenSyn model significantly improves all metrics. More importantly, Recall increases from 0.818 to 0.828, indicating that residual connections significantly enhanced the model's ability to identify positive examples. This result validates the critical role of residual connections in mitigating gradient vanishing and preserving fine-grained features in shallow layers within deep graph neural networks. When both GIN and residual connection were introduced simultaneously, the model achieved a peak AUC of 0.921 on the core metric while maintaining a favorable balance between PREC and Recall. 
Although the ablation experiments validated the effectiveness of key components, the analysis results also revealed several limitations warranting attention: Data analysis indicates that different components yield varying degrees of improvement across various metrics. Residual connections showed the most significant improvement in Recall, but their impact on PREC was relatively limited, whereas the GIN architecture performed better on the PREC metric. This ablation study primarily focuses on optimizing the feature extraction component, without conducting an in-depth analysis of the core attention interaction module. Future research should explore more sophisticated attention designs, such as multi-head attention and hierarchical attention, to further enhance model performance.
\begin{figure*}[!h]
    \centering
    \includegraphics[width=1\linewidth]{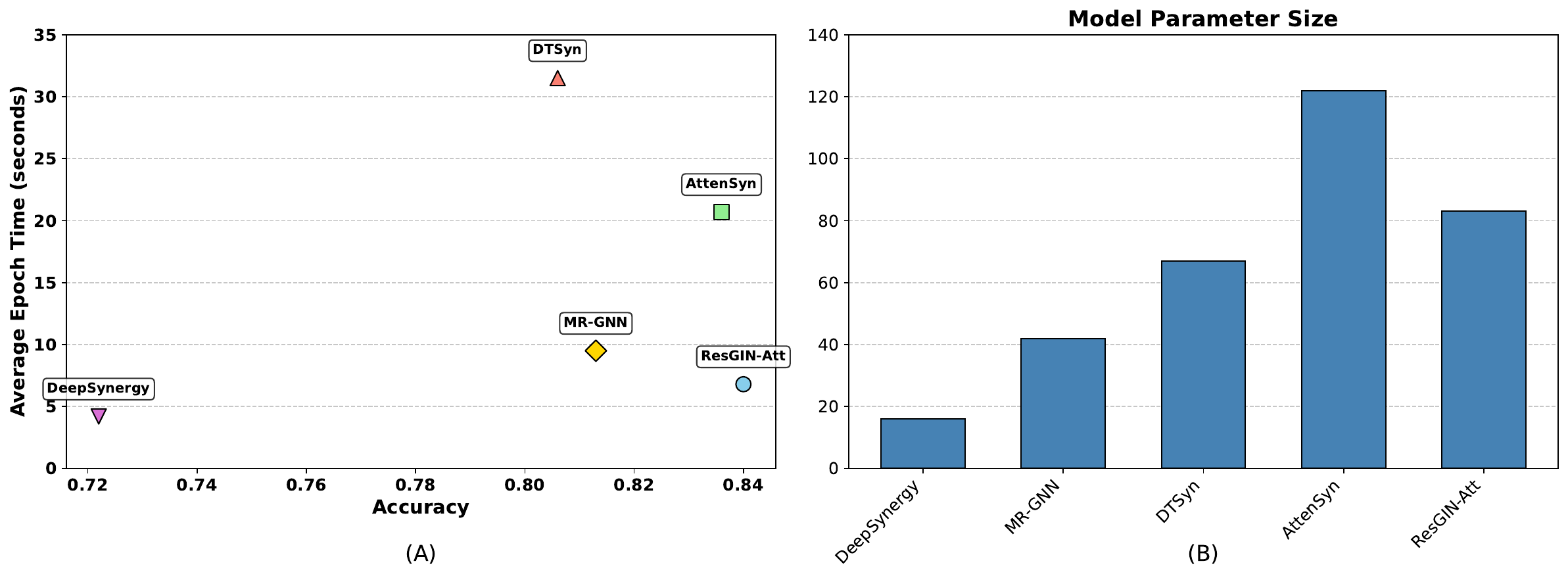}
    \caption{The each epoch times (A) and model parameter (B) between ResGIN-Att and baselines methods on O'Neil dataset.}
    \label{fig:model_comparison}
\end{figure*}
\subsection{Training time and model parameter}
In drug combination prediction, the execution time of the model is equally critical to predictive accuracy. Jointly, these two factors determine the model’s practical utility. Fig.\ref{fig:model_comparison} illustrates the relationship between the average training time per epoch and the accuracy of ResGIN-Att and several baseline methods. 

The results show that ResGIN-Att achieves high accuracy while maintaining a reasonable training time, significantly outperforming more complex models such as DTSyn, which requires nearly 35 seconds of training. In contrast, models such as DeepSynergy, while having shorter training times, exhibit significantly lower accuracy, highlighting the inherent trade-off between accuracy and computational efficiency in these approaches. By optimizing the balance between training time and prediction accuracy, ResGIN-Att demonstrates both efficiency and practicality in large-scale DDI prediction tasks. Although methods such as AttenSyn achieve excellent prediction accuracy, their lengthy training times further demonstrate that ResGIN-Att can effectively control computational costs while maintaining high predictive performance. 

Furthermore, as shown in the figure, ResGIN-Att has a relatively moderate parameter scale, whereas models such as AttenSyn require a significantly larger number of parameters. This further demonstrates the practical scalability of ResGIN-Att, ensuring that it maintains computational efficiency even when processing large scale datasets. The combination of architectural efficiency (residual connections, optimized aggregation) and parameter economy allows ResGIN-Att to deliver high predictive performance without incurring prohibitive computational costs. This makes it not only accurate but also practically deployable in real world pharmacological screening pipelines, where speed, precision, and scalability are equally essential.
\subsection{Parameter Sensitivity Analysis}
\subsubsection{The performance of different learning rate and dropout rate}
\begin{figure*}[t]
    \centering
    \includegraphics[width=1\linewidth]{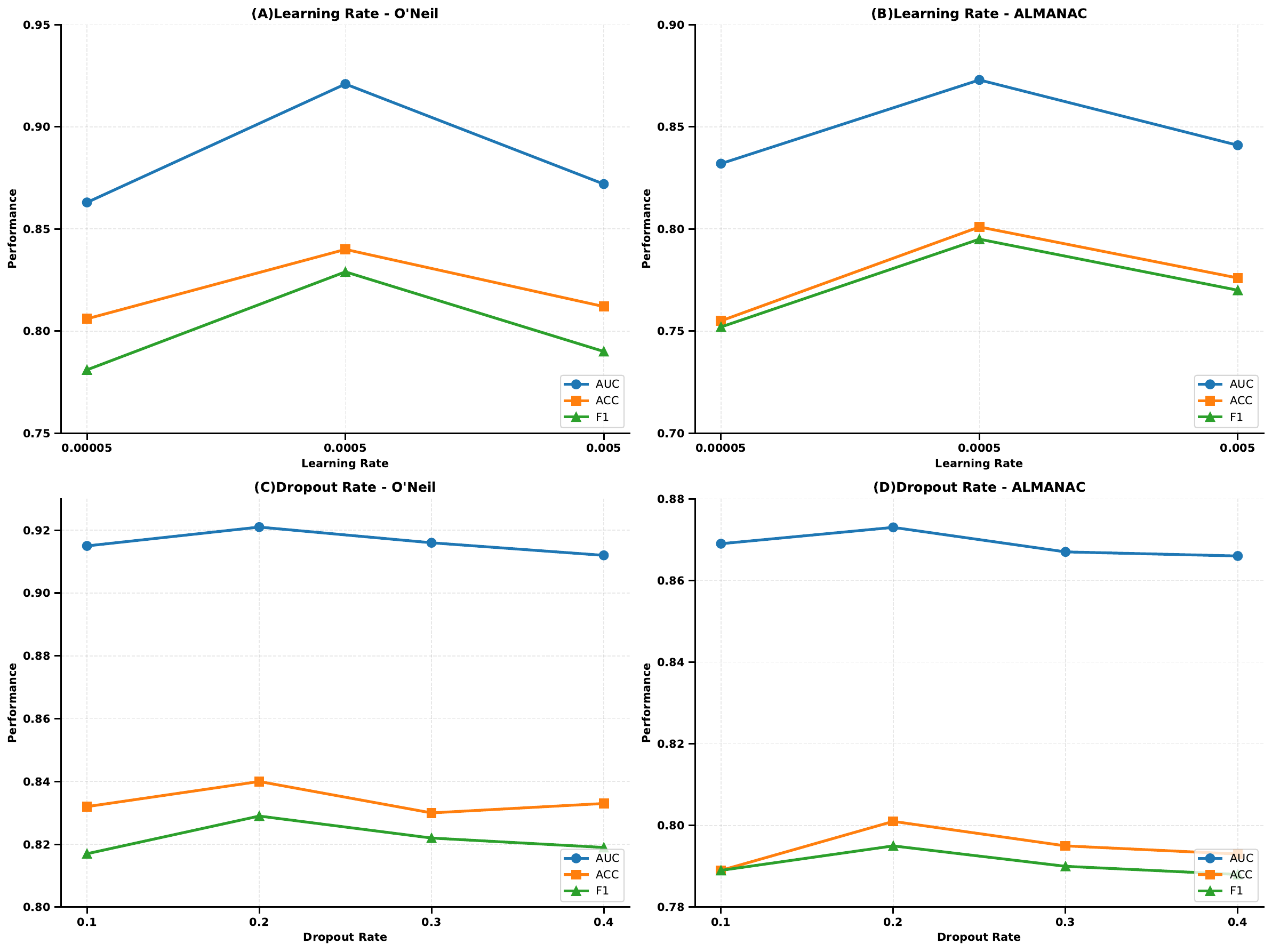}
    \caption{The performance of the different learning rate and dropout rate on the O’Neil and ALMANAC datasets.}
    \label{fig:learning}
\end{figure*}

To evaluate how sensitive the ResGIN-Att is to variations in learning rate and dropout rate, we conducted a systematic hyperparameter sweep in which the learning rate was set to 0.00005, 0.0005, and 0.005, and the dropout rate was set to 0.1, 0.2, 0.3, and 0.4. Fig.\ref{fig:learning} shows the model performance plots measured using four evaluation metrics on the O’Neil and ALMANAC datasets.
For the O’Neil dataset, the model performed best when the learning rate was 0.0005 and the dropout rate was 0.2. Similarly, for the ALMANAC dataset, optimal performance was also observed when the learning rate was 0.0005 and the dropout rate was 0.2. These results indicate that the model is highly sensitive to both the learning rate and the dropout rate, and optimal performance consistently occurs under these settings on both datasets. These findings suggest that although the O’Neil and ALMANAC datasets differ in graph density, they exhibit similar sensitivity to these parameters.
\subsubsection{The impact of different network depths}
\begin{figure}[h]
    \centering
    \includegraphics[width=1\linewidth]{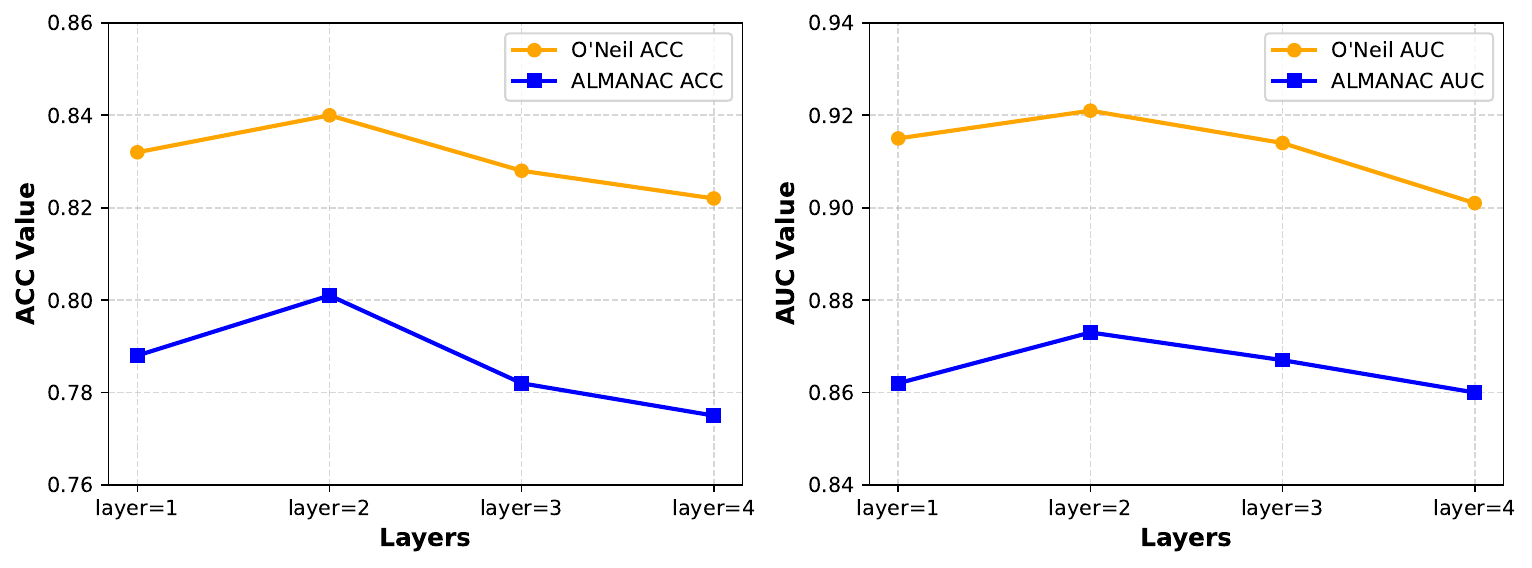}
    \caption{Effect of the different network depths on  O’Neil and ALMANAC dataset. Model performance initially improves as the network depth increases, but then declines. The model performs best when the network depth is 2.}
    \label{fig:Effect of the different network depths}
\end{figure}

To assess the sensitivity of the ResGIN-Att’s performance to the depth of the view encoder, we tested neural networks with [1, 2, 3, 4] layers, as shown in Fig.\ref{fig:Effect of the different network depths},
model performance peaks when the network depth is set to 2 layers. As the number of layers increases, performance gradually declines. Preliminary analysis suggests that as network depth increases, node features undergo iterative aggregation with those of neighboring nodes. Consequently, the representations of different nodes tend to converge, leading to the loss of individual features. This finding highlights the importance of selecting an optimal network depth, which involves balancing feature aggregation with the preservation of node-specific features to maximize model performance.

\subsection{Over-smoothing Analysis}
\begin{figure}[h]
    \centering
    \includegraphics[width=1\linewidth, height=0.5\linewidth]{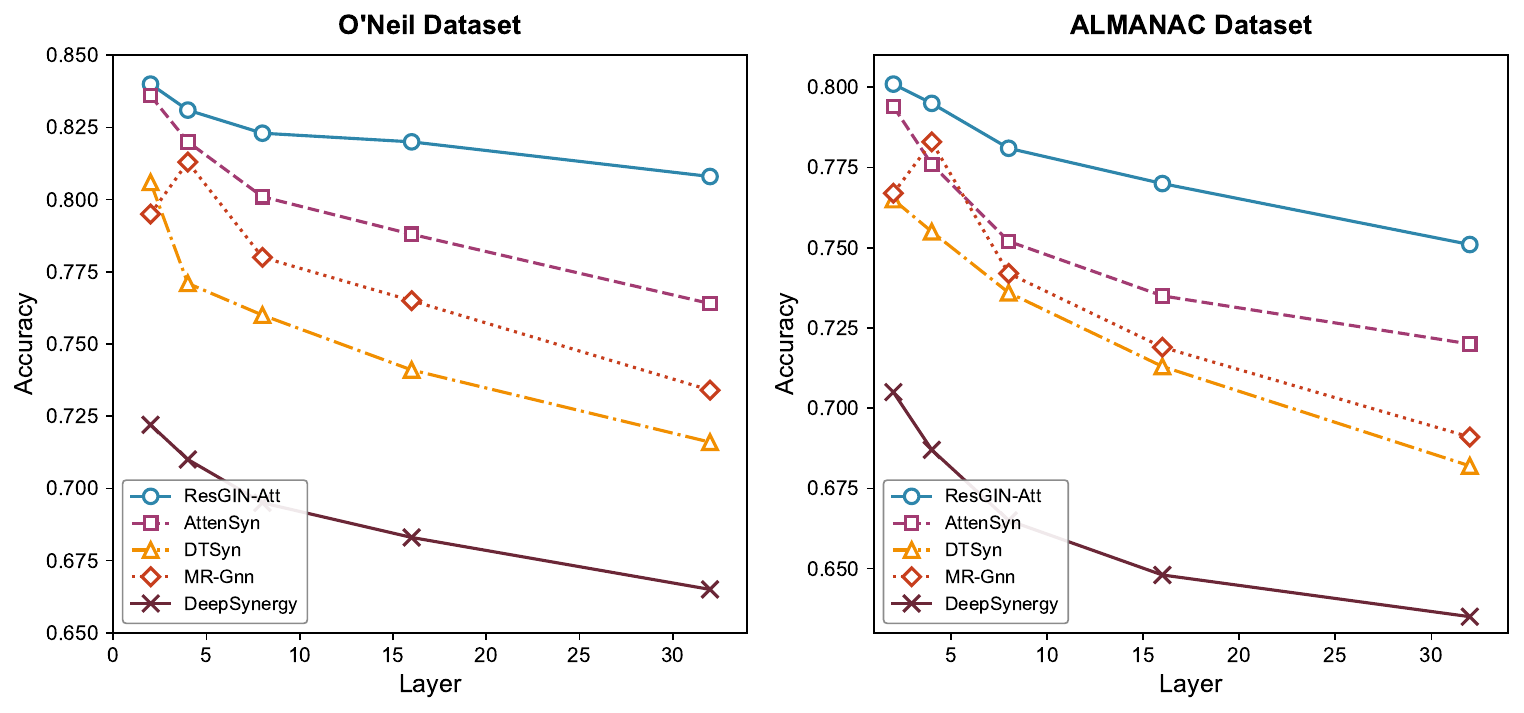}
    \caption{Performance analysis with varying network depths.}
    \label{fig:over_smoothing_analysis}
\end{figure}

The preliminary experiment above revealed that increasing depth may lead to performance degradation, a phenomenon closely related to the classic ``over-smoothing" problem in GNNs. Over-smoothing is a well-documented limitation in GNNs that constrains effective model depth. It arises during multi-layer message passing. As the number of layers increases, node representations progressively converge toward indistinguishable states. This erodes discriminative capacity. Consequently, performance degrades on downstream tasks such as classification and prediction. To validate the effectiveness of the residual mechanism introduced in the proposed ResGIN-Att model for mitigating over-smoothing, we systematically designed ``layer wise sensitivity" comparative experiments on two public drug interaction datasets: O'Neil and ALMANAC. The results are shown in Fig.\ref{fig:over_smoothing_analysis}.
The experiment compared the performance trajectories of ResGIN-Att against multiple mainstream baseline models (including AttenSyn, DTSyn, MR-GNN, DeepSynergy, etc.) by varying the network depth while keeping other hyperparameters consistent. The focus was on the patterns of accuracy (ACC) as depth increased.

Experimental results on the O’Neil dataset show that all benchmark models exhibit characteristic overfitting behavior: classification accuracy declines consistently and substantially with increasing network depth. While ResGIN-Att also follows the general pattern of ``slight performance degradation with increasing layers", its decline is markedly smaller than that of all benchmark models. More importantly, at each fixed layer count, ResGIN-Att maintained a comprehensive lead in accuracy over all benchmark models. This result preliminarily indicates that ResGIN-Att not only achieves superior performance but also exhibits lower sensitivity to depth increases. To enhance the universality and robustness of our conclusions, we replicated the aforementioned experiments on the ALMANAC dataset and observed identical patterns. ResGIN-Att again exhibited a smaller decline than the baseline model. The highly consistent experimental results across both datasets demonstrate ResGIN-Att's superior resilience against over-smoothing issues.

From the initial depth sensitivity analysis to the systematic, in-depth investigation of the over-smoothing problem, our experiments progress from observing the phenomenon to verifying the underlying mechanism. The results demonstrate that the ResGIN-Att model effectively mitigates over-smoothing across three complementary dimensions: its emergence, severity, and functional impact. Experimental data clearly demonstrate that compared to existing benchmark models, ResGIN-Att exhibits significantly lower sensitivity to increases in network depth and maintains performance advantages at any tested depth. This empirically demonstrates that residual connections are a key design element for enhancing the robustness and performance of deep graph neural networks. Not only do they strengthen the model's representational capabilities by preserving multi-level features, but they also provide an effective and concise solution to the long-standing over-smoothing problem in graph neural networks by establishing shortcut connections. This lays a solid foundation for constructing deeper and more powerful graph neural network models.
\section{Conclusion}
This paper has proposed ResGIN-Att, a novel graph neural network for drug synergy prediction. The core innovation has lain in integrating residual connections with graph isomorphism networks to mitigate over-smoothing, combined with a cross-attention mechanism to explicitly model drug-drug interactions. Experiments on five public datasets have demonstrated that ResGIN-Att achieves competitive performance against state-of-the-art baselines, while ablation, sensitivity, and over-smoothing analyses have collectively validated the rationality and robustness of the model design. Future work will focus on incorporating richer biological knowledge, such as protein-protein interaction networks and pathway data, to further enhance predictive accuracy and interpretability, ultimately supporting more effective combination therapy discovery.
\section{Conflicts of interest}
The authors have no conflicts of interest to declare.
\section{Data availability}
The code and data will be publicly accessible at \url{https://github.com/szerq/ResGIN-att} once this manuscript is accepted.
\section{Acknowledgments}
The work was supported by the Key Technologies for Integrated Planning and Operational Optimization of Power Systems in Industrial Parks to Promote the Integration of New Energy Sources (No. 2025YFA1017900) and National Undergraduate Training Program for Innovation and Entrepreneurship (No. 202510759062).
\bibliographystyle{unsrtnat} 
\bibliography{main.bbl}






\end{document}